\pdfoutput=1

\documentclass[11pt]{article}

\usepackage[final]{acl}

\usepackage{times}
\usepackage{latexsym}
\usepackage[T1]{fontenc}

\usepackage[utf8]{inputenc}
\usepackage{microtype}
\usepackage{inconsolata}
\usepackage{graphicx}

\usepackage{amsmath, amssymb}
\usepackage{tcolorbox}
\usepackage{longtable}
\usepackage{booktabs}
\usepackage{multirow}
\usepackage{afterpage}
\usepackage{enumitem}

\title{Interpreting the Effects of Quantization on LLMs}

\author{Manpreet Singh \\
  Dalhousie University \\
  \texttt{mn308829@dal.ca} \\\And
  Hassan Sajjad \\
  Dalhousie University \\
  \texttt{hsajjad@dal.ca} \\}

\begin{document}
\maketitle
\begin{abstract}
Quantization offers a practical solution to deploy LLMs in resource-constraint environments. However, its impact on internal representations remains understudied, raising questions about the reliability of quantized models. In this study, we employ a range of interpretability techniques to investigate how quantization affects model and neuron behavior. We analyze multiple LLMs under 4-bit and 8-bit quantization.
Our findings reveal that the impact of quantization on model calibration is generally minor. Analysis of neuron activations indicates that the number of dead neurons, i.e., those with activation values close to 0 across the dataset, remains consistent regardless of quantization. In terms of neuron contribution to predictions, we observe that smaller full precision models exhibit fewer salient neurons, whereas larger models tend to have more, with the exception of Llama-2-7B. The effect of quantization on neuron redundancy varies across models. Overall, our findings suggest that effect of quantization may vary by model and tasks, however, we did not observe any drastic change which may discourage the use of quantization as a reliable model compression technique. We make our code publicly available.\footnote{\url{https://github.com/MSingh-CSE/LLM-Quantization-Effects}}
\end{abstract}

\section{Introduction}
The last decade has seen a tremendous amount of work done in language modeling, specifically in large language models (LLMs) \cite{devlin2019bertpretrainingdeepbidirectional, Liu_2023, touvron2023llama2openfoundation}. There is a common trend to increase the number of parameters in LLMs to improve performance. However, this approach exacerbates the challenge of resource requirements, including computational and energy costs \cite{patterson2021carbonemissionslargeneural}. Quantization is a model compression technique that is widely used because of its effectiveness and simplicity \cite{bondarenko2024lowrankquantizationawaretrainingllms, dettmers2022llmint88bitmatrixmultiplication, wu2023understandingint4quantizationtransformer}. 
Quantization reduces the model size by using lower precision weights and/or activations, which can improve its inference speed while using less storage space. The effect of quantization is generally measured
by comparing a model's performance on downstream NLP tasks \cite{li2024evaluatingquantizedlargelanguage, neuralmagicquantization}. 

While performance on downstream tasks is crucial to understand the end-to-end impact, the evaluation is limited to a set of downstream tasks used for evaluation. In other words, it does not provide complete insights into the effect of quantization on the knowledge learned by models. In this work, we argue that the interpretation serves as an additional metric and evidence to analyze the effect of quantization on the model. 
For instance, it may reveal which types of knowledge or relationships are preserved or degraded by quantization, giving a deeper understanding of whether essential patterns remain intact. This is especially important for safety-critical applications such as finance, law, and healthcare \cite{hassan2024activelearningrobustrepresentative} where reliability of a model is necessary.

In this research, we study the effect of quantization, specifically LLMs quantized in 4-bit and 8-bit, to investigate its effect on the model's behavior and internal representations. To the best of our knowledge, this is first work that interpret the effect of quantization across various dimensions.
Specifically, to explore the behavior of the model and its neurons from multiple perspectives, we address the following key questions:

\begin{enumerate}[itemsep=-4pt, topsep=1pt]
    \item What is the effect of quantization on a model's confidence and calibration?
    \item Does quantization influence the contribution of neurons to model predictions? A major change in contributing neurons may reflect a change in model's decision making process.
    \item How does quantization affect the number of ``dead neurons''? A representation with a large number of dead neurons can cause the model to depend on only a handful of neurons.
    \item Does quantization affect the redundancy of neurons? In other words, does it result in more neurons learning identical information which may also reflect that some learned knowledge is lost during quantization.
\end{enumerate}

Broadly, our research questions aim to study the effect of quantization on the reliability of a model. For instance, an adverse affect on a model's confidence and calibration distorts output probabilities and may lead to unreliable uncertainty estimates in critical applications. Examining whether quantization alters the contribution of neurons to predictions provides insight into how compression reshapes neurons sensitivity to input features. Analyzing changes in the number of dead neurons helps assess whether quantization reduces network utilization or representational capacity, which can undermine robustness. Finally, exploring the redundancy of neurons i.e. whether quantization leads to more neurons learning similar information, offers valuable understanding of how model compression influences feature diversity and efficiency. Collectively, these objectives aim to reveal how quantization affects the representations and behavior of a model.

We analyze multiple open-source models, under two quantization settings: 4-bit \cite{dettmers2023qloraefficientfinetuningquantized} and 8-bit \cite{dettmers2022llmint88bitmatrixmultiplication} and 
compare them with the full-precision float-16 weight model. 

Our findings indicate that, while quantization does not cause drastic changes, its effects can vary depending on the specific context. A dataset and model interpretation might be necessary for reliably assessing the impact of quantization in practical settings.
We summarize our notable findings as follows:
\begin{enumerate}[itemsep=-4pt, topsep=1pt]
    \item Quantization does not lead to any substantial change in model confidence and calibration. 
    \item Based on neuron activations, quantization does not have a major effect, i.e., the number of dead neurons remains largely unchanged.
    \item Attribution-based analysis shows that full-precision models have fewer salient neurons in smaller LLMs and more in larger ones.
    \item Neuron redundancy differs between the subject models. In Phi-2, the full-precision model exhibits a higher number of correlated neuron pairs, indicating greater redundancy, whereas in Llama-2-7B, quantization causes only a minor difference in redundancy.
\end{enumerate}

\section{Methodology}
We study the model confidence and calibration, neuron activations, redundancy and attributions with respect to quantization.

\subsection{Confidence Analysis}
Confidence analysis aims to find the average confidence of a model in its predictions \cite{10.1016/j.inffus.2021.05.008}.
We calculate the average confidence of the model using the following equation:

\[
\text{Average Confidence} = \frac{1}{N} \sum_{i=1}^N \max  P(y_i)
\]

Here, \( N \) is the total number of data points in the dataset, and \( P(y_i) \) represents the softmax probability of the output label \( y_i \) with the highest probability for the \( i \)-th prediction. The term \( \max \big( P(y_i) \big) \) indicates the confidence of the model in its selected prediction for each datapoint.

\subsection{Calibration Analysis}
Calibration can be defined as the degree to which a model’s predicted probabilities reflect the actual frequencies of those outcomes \cite{nixon2020measuringcalibrationdeeplearning}. Despite high accuracy, deep neural networks often suffer from \textit{miscalibration} \cite{guo2017calibrationmodernneuralnetworks}.

We use the Adaptive Calibration Error (ACE) metric \cite{nixon2020measuringcalibrationdeeplearning}, which adjusts its assessment based on the actual distribution of confidence values, enabling a more flexible and precise evaluation of calibration. ACE is calculated as follows:

\[
\text{ACE} = \frac{1}{KR} \sum_{k=1}^{K} \sum_{r=1}^{R} \left| \text{acc}(r, k) - \text{conf}(r, k) \right|
\]

Here, \( K \) is the number of classes., \( R \) is the number of adaptive calibration ranges, \( \text{acc}(r, k) \) and \( \text{conf}(r, k) \) are the accuracy and confidence values for the adaptive range \( r \) for class \( k \), respectively. The calibration range \( r \) is determined by dividing the predictions into \( R \) equally populated intervals based on sorted confidence scores. This way, each range contains approximately \(\lfloor N/R \rfloor\) predictions, where \( N \) is the total number of data points.

\subsection{Neuron's Attribution } \label{salient_neurons}
A neuron's attribution refers to its role and significance in a model's predictions, as determined by attribution methods such as integrated gradient (IG) \cite{sundararajan2017axiomatic}. To evaluate the impact of quantization on neuron attributions, we analyze the number of salient neurons that contribute significantly to the model's predictions. This analysis shows quantization effects on the model's ability to identify and rely on the important features.

Using Layer IG, we obtain attribution scores for each input token for a given layer as:
\[
\text{IG}([x_1, x_2, \dots, x_n]) = \{a_1, a_2, \dots, a_n\}
\]

Here, \( x_i \) represents each input token and \( a_i \) is the attribution score for the token \( x_i \).

The attribution score \( a_i \) is calculated as the sum of the contributions from neurons in a layer as:

\[
a_i = \sum_{j=1}^{N} n_j
\]

where \( N \) is the total neurons in the given layer, and \( n_j \) is the attribution score of neuron \( j \).

\paragraph{Selection of Top Contributing Neurons:}
The input to the model consists of a sequence of tokens. We propose two separate methods to select the salient neuron with respect to the prediction. Specifically, we select most salient neurons based on 1) the most salient input token and 2) the input sequence and combine them. 
Each technique highlights neurons with varying levels of granularity and context sensitivity.

\textbf{Most attributed token-based:} In this technique, we only consider the most attributed token's (i.e., input token with max attribution score) representation and select neurons that have a normalized attribution score \(>0.7\). This identifies neurons that are most important in determining the model’s predictions for the specific context of the selected token.
Given as:
\[
x_{best} = \arg\max_i \{a_i\}
\]
\[
n_j^{\text{salient}} = \{n_j \mid \frac{n_j}{\max(n_j)} > 0.7 \}, \forall j \in \text{Layer}
\]
Here, \(a_i\) is the attribution score for token \(x_i\) and \(n_j\) is the attribution score of neuron \(j\) for \(x_{best}\).

\textbf{Input sequence-based:} To identify neurons that are salient in the context of the input sequence, we calculate the total attribution over the entire input sequence by summing the attributions across all input tokens. We select the neurons that have an attribution score \(>0.7\) after normalization. This approach ensures that the selected neurons reflect their contributions to the overall meaning of the input, rather than being limited to the most attributed token only.  Given as:
\[
s_j = \sum_{i=1}^{n} a_{ij}
\]
\[
n_j^{\text{salient}} = \{n_j \mid \frac{s_j}{\max(s_j)} > 0.7 \}, \forall j \in {Layer}
\]
Here, \(a_{ij}\) is the attribution of neuron \(j\) for token \(x_i\), and \(s_j\) is the total attribution score of neuron \(j\) summed over all tokens.

\textbf{Token-agnostic:} Here, we select the attribution score of a neuron based on its maximum attribution over all tokens in the input sequence. This selection emphasizes neurons important for any part of the input sequence,  regardless of specific tokens. Given as:
\[
m_j = \max_{i} \{a_{ij}\}
\]
\[
n_j^{\text{salient}} = \{n_j \mid \frac{m_j}{\max(m_j)} > 0.7 \}, \forall j \in {Layer}
\]
Here, \(a_{ij}\) is the attribution score of neuron \(j\) for token \(x_i\), and \(m_j\) is the maximum attribution score for neuron \(j\) over all tokens.

Using all the strategies outlined above, we identify the most important neurons contributing to a single datapoint prediction and collate it over the dataset. Although the same neurons may be selected under different strategies, we consider only one occurrence of each selected neuron.

\subsection{Neuron's Activations}
Since quantization reduces weight precision, it may increase the number of insignificant neurons. To identify them, we follow \citet{voita2023neuronslargelanguagemodels}, defining \textit{dead neurons} as those whose activations remain consistently near zero across the dataset.

\subsubsection{Dead/Insignificant Neurons}
\citet{voita2023neuronslargelanguagemodels} observed that the number of dead neurons increases with the growth of a model's size. Their analysis of the OPT language model family, which uses the ReLU activation function, shows that over 70\% of neurons in some layers are dead. We hypothesize that quantization, by reducing the precision of weights, may contribute to an increase in the number of dead neurons in the network.

Apart from ReLU, other activation functions such as GELU \cite{DBLP:journals/corr/HendrycksG16} and SiLU \cite{DBLP:journals/corr/ElfwingUD17} may not produce activation values that are exactly zero. To generalize the concept of dead neurons for these activation functions, we define a threshold of \(-0.1\) to \(0.1\), categorizing neurons as dead if their activation values consistently remain within this range across the dataset.
For different activation functions, we define dead neurons as follows:

\begin{equation}
\begin{split}
    n^{dead}_{j} (ReLU) = \{ n_j \mid &\quad a_{j,d} = 0, \\
    &\quad \forall d \in {dataset} \}
\end{split}
\end{equation}

\begin{equation}
\begin{split}
    n^{dead}_{j} (Other Activations) = \{ n_j \mid &\quad \\
    -0.1 \leq a_{j,d} \leq 0.1, \\
    \quad \forall d \in {dataset} \}
\end{split}
\end{equation}

Here, \( a_{j,d} \) represents the activation of neuron \( n_j \) for a given data point \( d \) in the dataset.

\subsection{Correlation Analysis} \label{correlation}
We hypothesize that a low-precision quantization may cause more neurons to represent identical information, i.e., as precision is reduced, high precision neuron values may map to the same low precision value. 
Similar to~\citet{DBLP:journals/corr/abs-2004-04010}, we calculate the Pearson correlation of neurons at a layer to identify neurons representing similar information. The Pearson correlation is given by:
\[
r = \frac{\sum_{i=1}^{n} (x_i - \mu_x)(y_i - \mu_y)}{\sqrt{\sum_{i=1}^{n} (x_i - \mu_x)^2} \cdot \sqrt{\sum_{i=1}^{n} (y_i - \mu_y)^2}}
\]
Here, \( x \) and \( y \) are activation arrays for the selected neuron pair. \( \mu_x \) and \( \mu_y \) are the means of \( x \) and \( y \), respectively, and \( n \) is the number of elements in the arrays. \( \sqrt{\sum_{i=1}^{n} (x_i - \mu_x)^2} \) and \( \sqrt{\sum_{i=1}^{n} (y_i - \mu_y)^2} \) are standard deviation for x and y respectively. The value of \( r \) ranges between -1 and 1, where \( r = 1 \) indicates perfect positive correlation, \( r = -1 \) indicates perfect negative correlation, and \( r = 0 \) indicates no linear correlation.

In this study, we use the absolute values of correlation to focus solely on the strength of the relationship. We consider a neuron pair to be redundant if their correlation score $r>0.8$.

\section{Experiment Setup}

\subsection{Datasets} \label{datasets}
We consider five datasets in this study: 
BoolQ \cite{clark2019boolqexploringsurprisingdifficulty}, the Jigsaw Toxicity dataset \cite{cjadams_2017}, Physical Interaction: Question Answering (PIQA) \cite{PiQA}, Hellaswag \cite{Hellaswag} and IMDB sentiment classification~\cite{sentiment_dataset}.

We select a random subset of the each dataset for our experiment. More specifically, we used 10k samples from a combination of train and validation sets of BoolQ, 9k samples from the Toxicity train set, 1,838 validation samples from PIQA, 5,000 validation samples from Hellaswag, and the IMDB training set. Instruction-tuned samples for datasets is available in Appendix \ref{sec:datasetsamples}.

These datasets test different capabilities of models: (1) question answering involving reading comprehension (BoolQ), (2) toxic language detection and social bias understanding (Toxicity), (3) physical commonsense reasoning (PIQA), (4) commonsense reasoning (Hellaswag), and (5) sentiment analysis and opinion understanding (IMDB).

\subsection{Models}
The primary models analyzed in our study are Phi-2 \cite{phi2}, Llama-2 7B \cite{touvron2023llama2openfoundation}, Qwen 2.5 3B and 7B \cite{qwen2025qwen25technicalreport}, and Mistral-7B \cite{jiang2023mistral7b}.

To examine the internal representations within these models, we focus on the output of the first feed-forward layer in the multi-layer perceptron (MLP) block, post-activation. We select this layer as our analysis on dead neurons expects output from the activation function. For computational efficiency, we conduct experiments using the first, middle and last decoder blocks of each model.

Since our subject models employ GELU (Phi-2) and SiLU (Llama, Qwen, Mistral) as activation functions, which do not produce exact zero activations,
we include the OPT-6.7B model from the OPT family \cite{zhang2022optopenpretrainedtransformer} to assess the behavior of ReLU activations for comparison. This model utilizes a decoder-only architecture similar to other subject models. 

During generation, the seed is set to 42, and default arguments from the Huggingface \texttt{transformers} library are used.

\begin{figure*}[]
    \centering
    \includegraphics[width=1\textwidth]{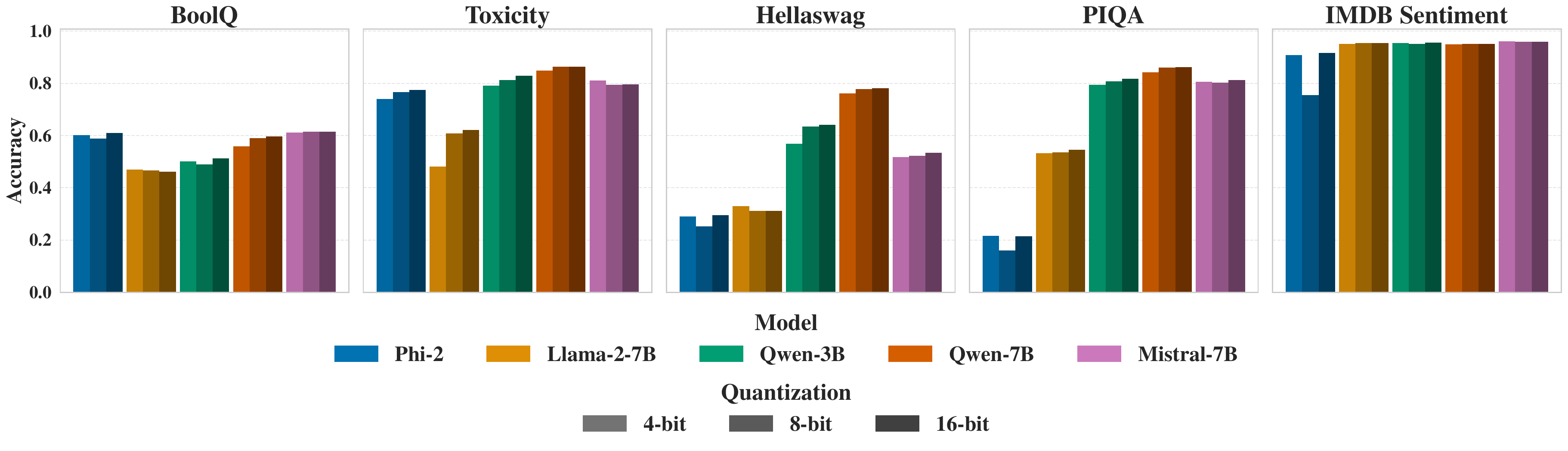} 
        \caption{Accuracy of subject models within different quantizations.} 
    \label{fig:accuracy}
\end{figure*}

\subsection{Quantization Configurations}
To perform comparative analysis across models under different quantization settings, we employed two widely-used quantization techniques: 4-bit \cite{dettmers2023qloraefficientfinetuningquantized} and 8-bit \cite{dettmers2022llmint88bitmatrixmultiplication}. Models are quantized using bitsandbytes config through Huggingface transformers. Table \ref{table:quantization} shows the hyperparameters used for quantization.

\begin{table}[]
    \centering
    \small
    \begin{tabular}{l|l}
        \toprule
        \textbf{Hyperparameter} & \textbf{Value} \\
        \multicolumn{2}{c}{\textbf{8-bit Quantization}} \\
        \midrule
        load\_in\_8bit & True \\
        bnb\_8bit\_compute\_dtype & torch.float16 \\
        bnb\_8bit\_use\_double\_quant & True \\
        \midrule
        \multicolumn{2}{c}{\textbf{4-bit Quantization}} \\
        load\_in\_4bit & True \\
        bnb\_4bit\_quant\_type & nf4 \\
        bnb\_4bit\_use\_double\_quant & True \\
        bnb\_4bit\_compute\_dtype & torch.float16 \\
        \bottomrule
    \end{tabular}
    \caption{Quantization Hyperparameters}
    \label{table:quantization}
    \vspace{-4mm}
\end{table}

\subsection{Attribution Technique}
We use Integrated Gradients \cite{sundararajan2017axiomatic} using Captum \cite{kokhlikyan2020captum} to find salient neurons in a  network.

\section{Findings}
In the following, we first report the accuracy of each model settings and then present our interpretation analysis.

\subsection{Accuracy}
We calculate accuracy to ensure that the models under observation have comparable performance under quantization. Since all the datasets require output to be a single word, we constrain model generation to a single token.

Figure \ref{fig:accuracy} presents a bar chart depicting the accuracy of subject models across various levels of quantization. The x-axis represents different quantization levels, while accuracy is displayed on the y-axis.
In most cases, quantization results in minimal degradation of model accuracy, typically within a range of 1–4\%. However, 4-bit quantization leads to substantial performance degradation for Llama-2-7B on the Toxicity (-14\%) and Qwen-3B on Hellaswag (-7\%). 
Similarly, the 8-bit quantized Phi-2 model shows reduced accuracy on PIQA (-5\%) and IMDB Sentiment (-17\%).

\subsection{Effect on Confidence and Calibration}
In this analysis, we observe the effect of quantization on the model's confidence and calibration. 

\subsubsection{Confidence Analysis}

\begin{figure*}[]
    \centering
    \includegraphics[width=1\linewidth]{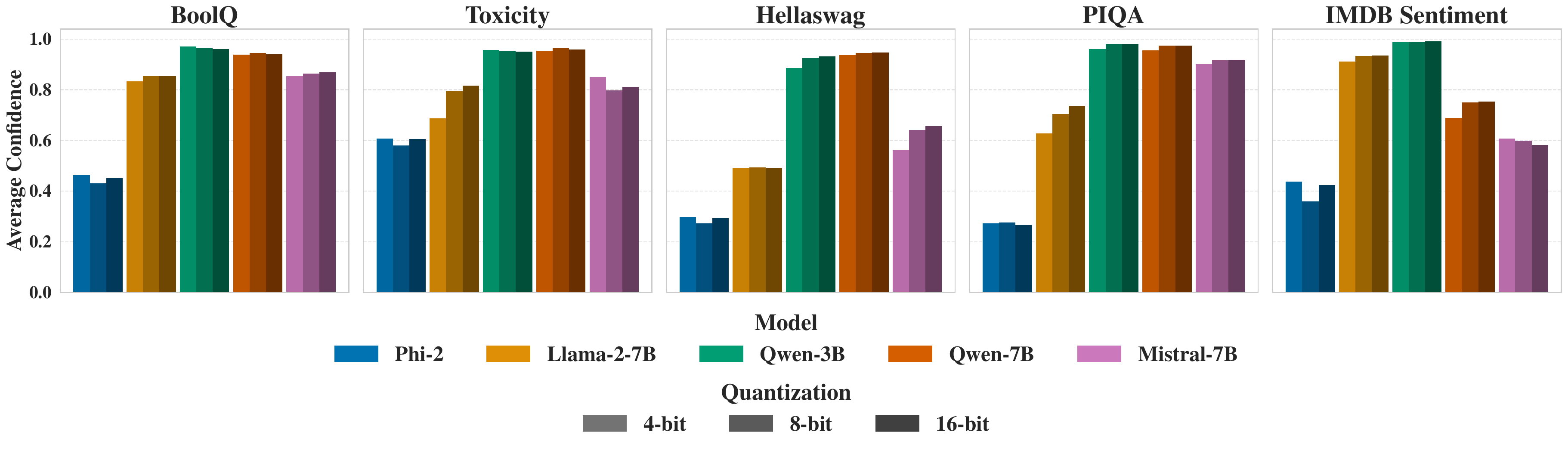}
    \caption{Average confidence of subject models under different quantizations.}
    \label{fig:average-confidence}
\end{figure*}
\begin{figure*}[]
    \centering
    \includegraphics[width=1\textwidth]{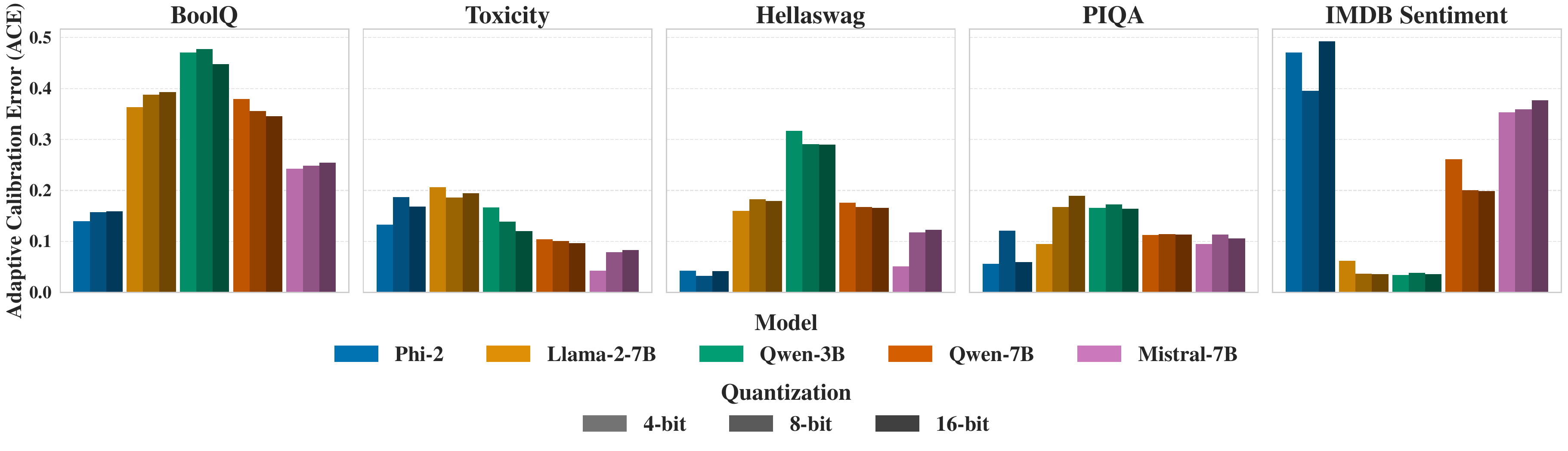 } 
    \caption{Adaptive Calibration Error (ACE) for subject models within different quantizations (lower is better).}
    \label{fig:ace_scores}
\end{figure*}

Figure \ref{fig:average-confidence} presents the average confidence of the evaluated models across various datasets. Broadly, the impact of quantization on model confidence appears limited, with only minor fluctuations observed. However, a trend emerges wherein 4-bit quantized models tend to exhibit slightly reduced confidence relative to their full-precision counterparts in most cases.

Notably, certain model-dataset pairs demonstrate more pronounced drops, suggesting that quantization may disproportionately affect specific tasks or models. For instance, the 4-bit quantized Llama-2-7B shows a reduction in confidence on the Toxicity and PIQA datasets, with decreases of 13\% and 11\%, respectively. Similarly, the 4-bit quantized Mistral-7B displays a 10\% confidence drop on Hellaswag, while the 4-bit quantized Qwen-7B shows a 6\% reduction on Sentiment Analysis.

These cases highlight the importance of task sensitivity when applying low-bit quantization. While average confidence remains relatively stable in general, targeted evaluation is essential to identify scenarios where confidence degradation may have downstream implications on reliability.

Interestingly, when comparing the average confidence to the corresponding accuracy results discussed earlier, we observe a notable disconnect: higher confidence does not consistently correlate with higher accuracy. This decoupling suggests that model confidence may not be a reliable proxy. Such a discrepancy motivates a deeper investigation into the calibration of these models, prompting our subsequent analysis on calibration to assess the alignment between confidence and correctness.

\subsubsection{Calibration Analysis}

Figure~\ref{fig:ace_scores} illustrates the Adaptive Calibration Error (ACE) for the evaluated models under varying levels of quantization. We observe that the effect of quantization on calibration is neither uniform across models nor consistent across datasets, indicating a strong dependency on both architectural and task specific factors.

For instance, 4-bit quantization leads to mixed outcomes for Llama-2-7b, 
where calibration error fluctuates, being higher for some tasks and lower for others—showing no clear pattern. In contrast, the Phi-2 model demonstrates more stable behavior under 4-bit quantization, with calibration error remaining similar or even improving in some cases. Interestingly, this pattern is reversed when models are quantized to 8-bit: Llama-2-7b exhibits consistently better calibration, whereas Phi-2 begins to show erratic changes in ACE across datasets.

Looking at the Qwen model family, both the 3B and 7B variants show increased or equivalent calibration error when quantized to lower bits, suggesting a reduced robustness in their confidence estimates under compression. Conversely, the Mistral models despite sharing same number of parameters with Qwen-7B, including same activation functions tend to exhibit improved calibration at lower bit.

The seemingly random fluctuations in ACE scores, particularly for certain model, dataset or weight precision combinations, could stem from several underlying factors. Differences in model pretraining objectives or tokenization strategies may contribute to how calibration responds to quantization.
Although quantization may introduce fluctuations in ACE, the difference is not substantial to undermine it's reliability, even often yielding improved calibration relative to full-precision variant.

\begin{table}[]
\renewcommand{\arraystretch}{1.3}
\centering
\footnotesize
\vspace{0.5em}
\begin{tabular}{l|c|c|c|c|r}
\toprule
\textbf{Model} & \textbf{Quant.} & \textbf{First} & \textbf{Mid.} & \textbf{Last} & \textbf{Total} \\ 
\midrule
\multirow{3}{*}{Phi-2} 
& 4-bit  & 69  & 1062 & 35  & 1166 \\ \cline{2-6}
& 8-bit  & 58  & 1034 & 43  & 1135 \\ \cline{2-6}
& 16-bit & 57  & 876  & 41  & 974  \\ \midrule
\multirow{3}{*}{Llama-2-7B} 
& 4-bit  & 34  & 1317 & 20  & 1371 \\ \cline{2-6}
& 8-bit  & 44  & 1256 & 17  & 1317 \\ \cline{2-6}
& 16-bit & 72  & 1191 & 18  & 1281 \\ \midrule
\multirow{3}{*}{Qwen-3B} 
& 4-bit  & 1283 & 3627 & 32 & 4942 \\ \cline{2-6}
& 8-bit  & 1104 & 3975 & 21 & 5100 \\ \cline{2-6}
& 16-bit & 960 & 3708 & 25 & 4693  \\ \midrule
\multirow{3}{*}{Qwen-7B} 
& 4-bit  & 700 & 3036 & 29 & 3765 \\ \cline{2-6}
& 8-bit  & 439 & 3394 & 45 & 3878 \\ \cline{2-6}
& 16-bit & 816 & 3261 & 34 & 4111 \\ \midrule
\multirow{3}{*}{Mistral-7B} 
& 4-bit  & 142 & 951 & 20 & 1113 \\ \cline{2-6}
& 8-bit  & 444 & 993 & 42 & 1479 \\ \cline{2-6}
& 16-bit & 513 & 936 & 38 & 1487 \\ 
\bottomrule
\end{tabular}
\caption{Number of salient neurons for subject models across quantizations (Quant.) within different layers.} \label{table:salient_neurons}
\vspace{-5mm}
\end{table}
\subsection{Effect on the Contribution of Neurons to Model Predictions}

Table \ref{table:salient_neurons} shows the count of salient neurons for subject models within different quantization, divided by layers. Given the high computational cost associated with computing attributions, we restricted this experiment to the BoolQ dataset.

We observe distinct trends in the number of salient neurons across quantization and model sizes. For smaller models such as Phi-2 and Qwen-3B, the full-precision model have fewer salient neurons compared to their quantized counterparts. This suggests that in these models, full precision enables more generalized neurons, where only a subset of neurons significantly contribute to the final prediction. In contrast, quantization introduces perturbations, likely increasing representational noise affecting generalization. As a result, more neurons become involved in the prediction process, compensating for the reduced expressivity of neurons.

This trend is reversed for some larger models. In Qwen-7B and Mistral-7B, we observe more salient neurons in the full-precision compared to the quantized variant. This may reflect the ability of larger models in full precision to utilize richer, more distributed representations, which are partially suppressed or sparsified under quantization.

Interestingly, Llama-2-7B does not follow the trend and aligns more closely with smaller models such as Phi-2. It has fewer salient neurons in full precision than in its 4-bit quantized version but similar to 8-bit. This divergence may stem from architectural differences, particularly in hidden layer size as 11,008 (same as Qwen-3B), compared to 18,944 in Qwen-7B and 14,336 in Mistral-7B.

\begin{table}[]
\renewcommand{\arraystretch}{1.3}
\centering
\footnotesize
\begin{tabular}{l|c|c|c|c}
\toprule
\textbf{Model} & \textbf{Quant.} & \textbf{F (\%)} & \textbf{M (\%)} & \textbf{L (\%)} \\
\midrule
\multirow{ 3}{*}{OPT-6.7B}      & 4-bit  & 23.43  & 0.35  & \textbf{0.12}  \\ \cline{2-5}
                                & 8-bit  & 23.45  & 0.26  & 0.15  \\ \cline{2-5}
                                & 16-bit & \textbf{23.35}  & \textbf{0.24}  & 0.14  \\ 
                                \midrule
\multirow{ 3}{*}{Phi-2}         & 4-bit  & \textbf{21.46}  & 0.00  & 0.01  \\ \cline{2-5}
                                & 8-bit  & 21.52  & 0.00  & 0.01  \\ \cline{2-5}
                                & 16-bit & 21.51  & 0.00  & 0.01  \\
                                \midrule
\multirow{ 3}{*}{Llama-2-7B}    & 4-bit  & 0.05   & 0.00  & 0.00  \\ \cline{2-5}
                                & 8-bit  & \textbf{0.04}   & 0.00  & 0.00  \\ \cline{2-5}
                                & 16-bit & 0.05   & 0.00  & 0.00  \\ 
                                \midrule
\multirow{ 3}{*}{Qwen-3B \& 7B}       & 4-bit  & 0.00   & 0.00  & 0.00  \\ \cline{2-5}
                                & 8-bit  & 0.00   & 0.00  & 0.00  \\ \cline{2-5}
                                & 16-bit & 0.00   & 0.00  & 0.00  \\ 
                                \midrule
\multirow{ 3}{*}{Mistral-7B}    & 4-bit  & 0.02   & 0.00  & 0.00  \\ \cline{2-5}
                                & 8-bit  & \textbf{0.01}   & 0.00  & 0.00  \\ \cline{2-5}
                                & 16-bit & 0.02   & 0.00  & 0.00  \\ 
                                \bottomrule
\end{tabular}
\caption{Percentage of dead neurons across models and quantizations (Quant.) within different layers (F: First, M: Middle, L: Last).}
\label{table:dead_neurons}
\end{table}

Overall, the number of salient neurons serves as a proxy for how distributed or localized the decision-making process is within the network \cite{durrani2024discoveringsalientneuronsdeep}. Full precision models tend to use fewer, neurons when they are smaller. In contrast, in larger models, full precision can enable richer and more distributed neuron contribution.

\subsection{Effect on the number of ``dead neurons''}

As shown in Table \ref{table:dead_neurons}, quantization causes only a minor change in the count of dead neurons. The trend across quantization seems to be consistent, as the number of dead neurons remains almost similar between quantized and full-precision models.  

The pattern of higher neurons in initial layer in Phi-2 and OPT-6.7B likely reflects the role of initial layers in learning sparse, low-level features, while later layers capture higher-level contextual features \cite{dalvi2022discoveringlatentconceptslearned, voita2023neuronslargelanguagemodels}. We hypothesize that the consistently low count of dead neurons among Llama, Qwen and Mistral is due to the use of the SiLU activation function.

\begin{figure*}[h]
    \centering
    \includegraphics[width=1\linewidth]{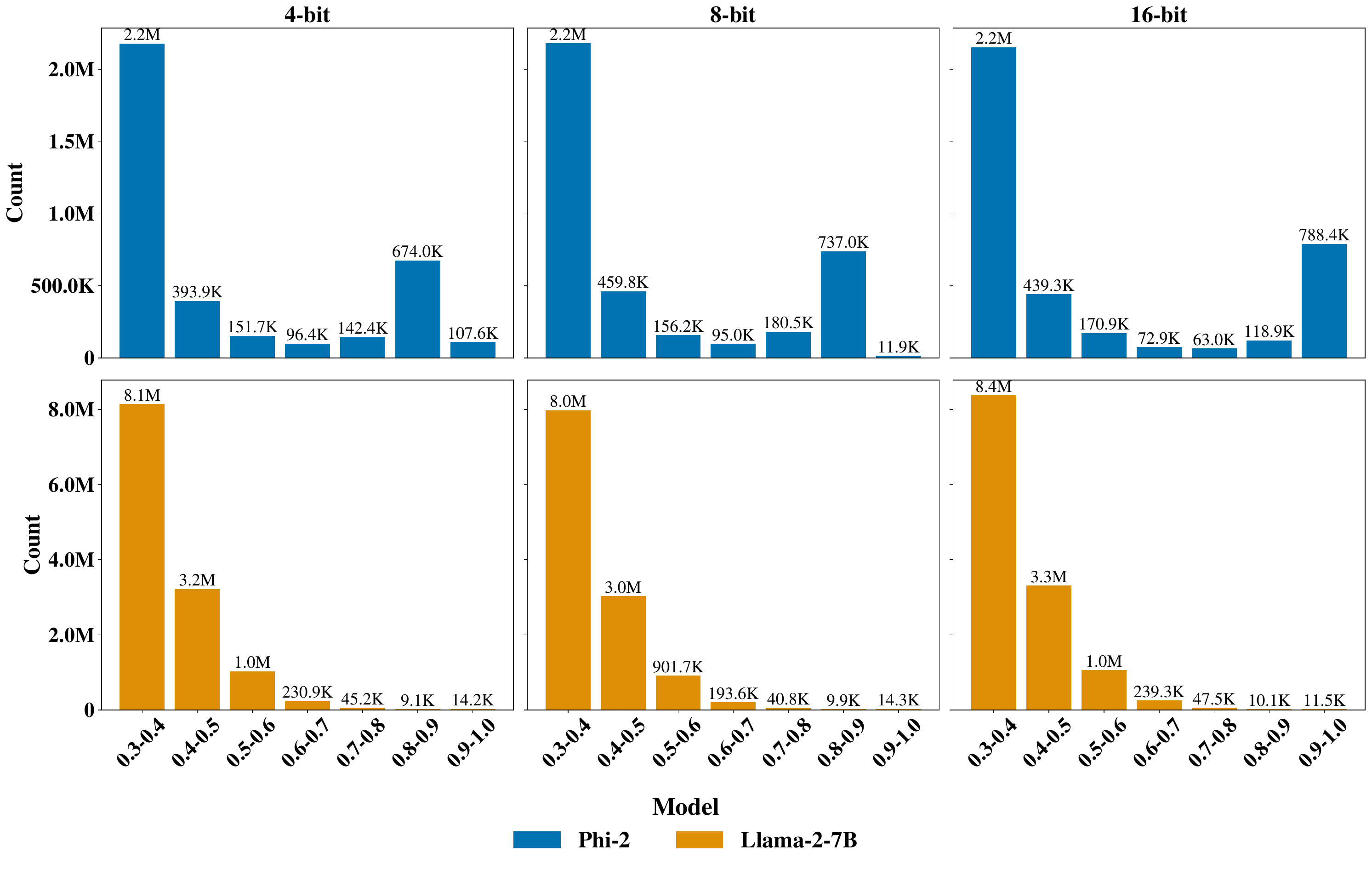}
    \caption{Neurons pair count based on correlation for Phi-2 and Llama-2-7B.}
    \label{fig:correlation_combined}
\end{figure*}

\subsection{Effect on the Redundancy of Neurons}
As identified in the works of \citet{DBLP:journals/corr/abs-2004-04010} language models can maintain 97\% of accuracy while using only 10\% of the neurons. This finding is valuable for model pruning. We investigate whether quantization leads to higher redundancy. Due to the substantial computational requirements, our analysis was limited to the Phi-2 and Llama-2-7B models, using activations from BoolQ.

\subsubsection{Correlation Analysis}

Figures \ref{fig:correlation_combined} shows neuron pairs count corresponding to correlation scores for 4-bit, 8-bit and full-precision variants of Phi-2 and Llama-2-7B. The X-axis highlights the different correlation score bins ranging from 0.3-0.4 to 0.9-1.0. This binning process helps to clearly observe the redundant neuron pairs count across all the layers. The Y-axis shows the count of neuron pairs that fall in that bin. Notice that the count is given for neuron pairs across all the layers, as our main focus is to observe the effect on redundancy of neurons within quantizations. For clarity in visualizing highly correlated neuron pairs, we excluded the 0.0–0.1, 0.1–0.2, and 0.2–0.3 correlation bins from the bar graph, since these bins contained similar numbers of uncorrelated neurons across different quantizations.

Considering highly correlated neurons, i.e., bins having correlation score \textgreater=0.8, Phi-2 in full precision shows the highest redundancy, with 907,352 correlated neuron pairs, compared to 781,583 in the 4-bit and 748,867 in the 8-bit configurations. This points to Phi-2 in full-precision having higher redundancy compared to quantized models.

In Llama-2-7B, the 8-bit model has the highest redundancy with 24,124 correlated neuron pairs, which is slightly better in 4-bit with 23,315 pairs. unlike Phi-2, the full-precision Llama-2-7B has the fewest correlated pairs (21,644), indicating lower redundancy compared to its quantized versions. However, the difference between neuron pairs in quantized versions is not as substantial as Phi-2.

We hypothesize that the difference in redundancy between Phi-2 and Llama-2-7B is likely due to difference in the number of dead neurons (Table \ref{table:dead_neurons}). For instance, in the initial layer, Phi-2 has over 20\% dead neurons, whereas Llama-2-7B exhibits nearly none. This variation in activations between the two models can likely be attributed to their different activation functions i.e. GELU in Phi-2 and SiLU in Llama-2-7B.

\section{Related Work}
This section reviews the relevant literature in quantization techniques and their analysis.

\textbf{Quantization Techniques}. Quantization \cite{720541} is used to reduce the memory requirement by reducing the size of weight and/or activation and increasing the inference time of a model \cite{jacob2017quantizationtrainingneuralnetworks, DBLP:journals/corr/abs-2103-13630}. 

Quantization-aware training (QAT) is costly and uses re-training of a model on a dataset to maintain accuracy  \cite{liu2023llmqatdatafreequantizationaware, du2024bitdistillerunleashingpotentialsub4bit, dettmers2023qloraefficientfinetuningquantized, kim2023memoryefficientfinetuningcompressedlarge}.

Post-training quantization quantizes models without any additional finetuning of the model with a limited dataset, but also suffers from performance issues \cite{banner2019posttraining4bitquantizationconvolution, DBLP:journals/corr/abs-2001-00281}. In case of LLM's Post Training Quantization can be of 3 types: i) Weight-Only Quantization \cite{park2024lutgemmquantizedmatrixmultiplication, frantar2023gptqaccurateposttrainingquantization, chee2024quip2bitquantizationlarge, lin2024awqactivationawareweightquantization}, ii) Weight-Activation Quantization \cite{NEURIPS2022_adf7fa39, yuan2023rptqreorderbasedposttrainingquantization, 10.1145/3579371.3589038, wei-etal-2023-outlier}, and iii) KV Cache Quantization \cite{hooper2024kvquant10millioncontext, yue2024wkvquantquantizingweightkeyvalue}.

\textbf{Quantization Analysis and Interpretation}. 
A number of works interpret models in their ability to learn language  phenomenon such as morphology~\cite{belinkov-etal-2017-neural} and syntax~\cite{arps-etal-2022-probing,hall-maudslay-cotterell-2021-syntactic,hupkes_diagnostic}. These works often do not relate the representation of linguistic phenomenon to end performance of the model~\cite{probing_shortcomings}. In this work, we aim to experiment with a diverse set of interpretation methods which can be correlated to the performance of the model.

\citet{DBLP:journals/corr/abs-2111-08163} explores confidence and calibration relation between quantized and full-precision model by using symmetric quantization. \citet{proskurina-etal-2024-quantization} shows quantization improves calibration in LLMs using GPTQ. Some literature exlores interpretation withing quantized model for vision model \cite{Norrenbrock_Rudolph_Rosenhahn_2024, Xpression, QuantizedAndInterpretable, rezabeyk2024saliencyassistedquantizationneural, 10755934}.

\section{Conclusion}
In this study, we have investigated the impact of quantization on internal representations of LLMs.
Confidence and Calibration analysis reveal that calibration remains mostly stable across quantization. Neuron's attributions highlights even while number of salient neurons change with quantization i.e. effect is reversed for smaller models and larger models, the quantization seems to maintain the generalization ability of neurons. In terms of activations, there is no major change in number of dead neurons. In terms of redundancy, Phi-2 and Llama-2-7B exhibit different patterns. As in the case of Phi-2 in full-precision had a higher number of neurons learning similar information, while in Llama-2-7B, there was a minor difference between highly correlation neuron pairs.

The effect of quantization vary across datasets. A dataset level interpretation is often needed to reliably measure the effect of quantization.

Overall, the results suggest that the effect of quantization could be dependent on the task and model's architecture. However, we don't see any major effect that could discourage the use of quantization as a reliable approach for model deployment.

\section{Limitations}
This study has certain limitations that should be considered when interpreting the results. Due to computational constraints, our experiments were limited to specific quantization configurations, model sizes, and datasets, which may not fully capture the impact of quantization across all LLMs or in varied deployment settings. Extreme quantizations such as 2-bit and 3-bit can be added to explore the effects within these quantizations.
Currently we investigated with tasks which required single token output, generative tasks such as coding, summarization etc. can be explored.

\section*{Acknowledgments}
We acknowledge the support of the Natural Sciences and Engineering Research Council of Canada (NSERC), RGPIN-2022-03943, Canada Foundation of Innovation (CFI) and Research Nova Scotia. Advanced computing resources are provided by ACENET, the regional partner in Atlantic Canada, and the Digital Research Alliance of Canada.  

\bibliography{custom}

\appendix

\section{Dataset Samples}
\label{sec:datasetsamples}
Table \ref{table:boolq_samples} shows samples from BoolQ. 
Table \ref{table:jigsaw_samples} shows samples from Toxicity.
Table \ref{table:piqa_samples} shows samples from PIQA.
Table \ref{table:hellaswag_samples} shows samples from Hellaswag.
Table \ref{table:sentiment_samples} shows samples from IMDB sentiment.

\section{Analysis of Salient and Dead Neurons Across Thresholds}
To further examine the neuron behavior, we analyze both salient and dead neurons across multiple thresholds. Table \ref{table:salient_neurons_thresholds} shows salient neurons count within different threshold and Table \ref{table:dead_neurons_thresholds} shows overall dead neurons percentage within different thresholds, the absolute number of neurons identified in each category changes with the threshold choice; however, the overall trend remains consistent. This indicates that while quantization and threshold sensitivity affect the magnitude of neuron counts, the patterns are robust across threshold settings.

\begin{table}[h]
\centering
\footnotesize
\setlength{\tabcolsep}{3pt}
\begin{tabular}{p{0.7\columnwidth} c}
\toprule
\textbf{Prompt} & \textbf{Gold} \\
\midrule
Respond with True or False only.\\
do iran and afghanistan speak the same language\\
Answer: & True \\ \midrule
Respond with True or False only.\\
do good samaritan laws protect those who help at an accident\\
Answer: & True \\ \midrule
Respond with True or False only.\\
is windows movie maker part of windows essentials\\
Answer: & True \\ \midrule
Respond with True or False only.\\
is confectionary sugar the same as powdered sugar\\
Answer: & True \\ \midrule
Respond with True or False only.\\
is elder scrolls online the same as skyrim\\
Answer: & False \\
\bottomrule
\end{tabular}
\caption{Samples from BoolQ dataset.}
\label{table:boolq_samples}
\end{table}

\begin{table}
    \centering
    \footnotesize
    \begin{tabular}{p{0.7\columnwidth} c}
        \toprule
        \textbf{Prompt} & \textbf{Gold} \\
        \midrule
        Given a sentence, respond with only `True' if the sentence is toxic or else `False'.\\
        Sentence: `Red herrings to continue to excuse Breivik. Go away, nasty person.'\\ Answer: & True \\
        \midrule
        Given a sentence, respond with only `True' if the sentence is toxic or else `False'.\\
        Sentence: `That is just plain BS, forgive the wording....if you look at that rule, it is meant for ``empty" sites that redirect to a site with content. The original URL is to a site with content. User:kalimantan'\\ Answer: & False \\
        \midrule
        Given a sentence, respond with only `True' if the sentence is toxic or else `False'.\\
        Sentence: `Iori Furukawa There's a name warrior at Iori Furukawa, thought you should check it out since you moved it. Thanks,'\\ Answer: & False \\
        \midrule
        Given a sentence, respond with only `True' if the sentence is toxic or else `False'.\\
        Sentence: `YOU'RE BLOCKED FOR 700 DAYS. SHUSHING COW'\\ Answer: & True \\
        \midrule
        Given a sentence, respond with only `True' if the sentence is toxic or else `False'.\\
        Sentence: `Bautista's Bat Flip Do you think it should be added in this article or in 2015 Toronto Blue Jays season article? If it should be added, then an image of the bat flip should be added as well. What do you think?'\\ Answer: & False \\
        \bottomrule
    \end{tabular}
    \caption{Samples from Jigsaw Toxicity dataset}
    \label{table:jigsaw_samples}
\end{table}

\begin{table}
    \centering
    \footnotesize
    \begin{tabular}{p{0.7\columnwidth} c}
        \toprule
        \textbf{Prompt} & \textbf{Gold} \\
        \midrule
        Respond with only the correct label (A or B) that best describes the appropriate steps for completing the task. Do not include any additional text or explanation, only respond with one letter.\\
        Task: How do I ready a guinea pig cage for it's new occupants?\\
        Options:\\
        A: Provide the guinea pig with a cage full of a few inches of bedding made of ripped paper strips, you will also need to supply it with a water bottle and a food dish.\\
        B: Provide the guinea pig with a cage full of a few inches of bedding made of ripped jeans material, you will also need to supply it with a water bottle and a food dish.\\
        Answer: & A \\
        \midrule
        Respond with only the correct label (A or B) that best describes the appropriate steps for completing the task. Do not include any additional text or explanation, only respond with one letter.\\
        Task: dresser\\
        Options:\\
        A: replace drawer with bobby pin\\
        B: finish, woodgrain with  bobby pin\\
        Answer: & B \\
        \midrule
        Respond with only the correct label (A or B) that best describes the appropriate steps for completing the task. Do not include any additional text or explanation, only respond with one letter.\\
        Task: To fight Ivan Drago in Rocky for sega master system.\\
        Options:\\
        A: Drago isn't in this game because it was released before Rocky IV.\\
        B: You have to defeat Apollo Creed and Clubber Lang first.\\
        Answer: & B \\
        \midrule
        Respond with only the correct label (A or B) that best describes the appropriate steps for completing the task. Do not include any additional text or explanation, only respond with one letter.\\
        Task: Make outdoor pillow.\\
        Options:\\
        A: Blow into tin can and tie with rubber band.\\
        B: Blow into trash bag and tie with rubber band.\\
        Answer: & B \\
        \midrule
        Respond with only the correct label (A or B) that best describes the appropriate steps for completing the task. Do not include any additional text or explanation, only respond with one letter.\\
        Task: ice box\\
        Options:\\
        A: will turn into a cooler if you add water to it\\
        B: will turn into a cooler if you add soda to it\\
        Answer: & A \\
        \bottomrule
        
    \end{tabular}
    \caption{Samples from PIQA}
    \label{table:piqa_samples}
\end{table}

\begin{table}
    \centering
    \footnotesize
    \begin{tabular}{p{0.7\columnwidth} c}
        \toprule
        \textbf{Prompt} & \textbf{Gold} \\
        \midrule
        A man is sitting on a roof. he\\
        Choose the most appropriate continuation:\\
        0. is using wrap to wrap a pair of skis.\\
        1. is ripping level tiles off.\\
        2. is holding a rubik's cube.\\
        3. starts pulling up roofing on a roof.\\
        Answer with only the number.\\
        Answer: & 3 \\
        \midrule
        A lady walks to a barbell. She bends down and grabs the pole. the lady\\
        Choose the most appropriate continuation:\\
        0. swings and lands in her arms.\\
        1. pulls the barbell forward.\\
        2. pulls a rope attached to the barbell.\\
        3. stands and lifts the weight over her head.\\
        Answer with only the number.\\
        Answer: & 3 \\
        \midrule
        Two women in a child are shown in a canoe while a man pulls the canoe while standing in the water, with other individuals visible in the background. the child and a different man\\
        Choose the most appropriate continuation:\\
        0. are then shown paddling down a river in a boat while a woman talks.\\
        1. are driving the canoe, they go down the river flowing side to side.\\
        2. sit in a canoe while the man paddles.\\
        3. walking go down the rapids, while the man in his helicopter almost falls and goes out of canoehood.\\
        Answer with only the number.\\
        Answer: & 2 \\
        \midrule
        A boy is running down a track. the boy\\
        Choose the most appropriate continuation:\\
        0. runs into a car.\\
        1. gets in a mat.\\
        2. lifts his body above the height of a pole.\\
        3. stands on his hands and springs.\\
        Answer with only the number.\\
        Answer: & 2 \\
        \midrule
        The boy lifts his body above the height of a pole. The boy lands on his back on to a red mat. the boy\\
        Choose the most appropriate continuation:\\
        0. turns his body around on the mat.\\
        1. gets up from the mat.\\
        2. continues to lift his body over the pole.\\
        3. wiggles out of the mat.\\
        Answer with only the number.\\
        Answer: & 1 \\
        \bottomrule
        
    \end{tabular}
    \caption{Samples from Hellaswag}
    \label{table:hellaswag_samples}
\end{table}

\begin{table}
    \centering
    \footnotesize
    \begin{tabular}{p{0.7\columnwidth} c}
        \toprule
        \textbf{Prompt} & \textbf{Gold} \\
        \midrule
        Review:\\
        A wonderful little production. The filming technique is very unassuming- very old-time-BBC fashion and gives a comforting, and sometimes discomforting, sense of realism to the entire piece. The actors are extremely well chosen- Michael Sheen not only "has got all the polari" ....\\
        What is the sentiment of this review? Answer with only one word: positive or negative.\\
        Answer: & positive \\
        \midrule
        Review:\\
        I thought this was a wonderful way to spend time on a too hot summer weekend, sitting in the air conditioned theater and watching a light-hearted comedy. The plot is simplistic, but the dialogue is witty and the characters are likable (even the well bread suspected serial killer). ...\\
        What is the sentiment of this review? Answer with only one word: positive or negative.\\
        Answer: & positive \\
        \midrule
        Review:\\
        Basically there's a family where a little boy (Jake) thinks there's a zombie in his closet \& his parents are fighting all the time. This movie is slower than a soap opera... and suddenly, Jake decides to become Rambo and kill the zombie. OK, first of all when you're going to....\\
        What is the sentiment of this review? Answer with only one word: positive or negative.\\
        Answer: & negative \\
        \midrule
        Review:\\
        Petter Mattei's "Love in the Time of Money" is a visually stunning film to watch. Mr. Mattei offers us a vivid portrait about human relations. This is a movie that seems to be telling us what money, power and success do to people in the different situations we encounter. This being a variation on....\\
        What is the sentiment of this review? Answer with only one word: positive or negative.\\
        Answer: & positive \\
        \midrule
        Review:\\
        Probably my all-time favorite movie, a story of selflessness, sacrifice and dedication to a noble cause, but it's not preachy or boring. It just never gets old, despite my having seen it some 15 or more times in the last 25 years....\\
        What is the sentiment of this review? Answer with only one word: positive or negative.\\
        Answer: & positive \\
        \bottomrule
        
    \end{tabular}
    \caption{Samples from IMDB Sentiment dataset}
    \label{table:sentiment_samples}
\end{table}

\begin{table}
\centering
\footnotesize
\setlength{\tabcolsep}{4pt}
\renewcommand{\arraystretch}{1.1}
\begin{tabular}{l l c c c}
\toprule
\textbf{Model} & \textbf{Quant.} & \textbf{Thr=0.7} & \textbf{Thr=0.8} & \textbf{Thr=0.9} \\
\midrule
Phi-2 & 4-bit  & 1166 & 867 & 679 \\
      & 8-bit  & 1135 & 847 & 676 \\
      & 16-bit &  974 & 721 & 547 \\ \midrule
Llama-2-7B & 4-bit  & 1371 & 1037 & 809 \\
            & 8-bit  & 1317 &  996 & 756 \\
            & 16-bit & 1281 &  957 & 732 \\ \midrule
Qwen-3B & 4-bit  & 4942 & 4075 & 3369 \\
         & 8-bit  & 5100 & 4198 & 3549 \\
         & 16-bit & 4693 & 3887 & 3240 \\ \midrule
Qwen-7B & 4-bit  & 3765 & 2923 & 2326 \\
         & 8-bit  & 3878 & 3016 & 2421 \\
         & 16-bit & 4111 & 3220 & 2622 \\ \midrule
Mistral-7B & 4-bit  & 1113 &  871 & 672 \\ 
            & 8-bit  & 1479 & 1165 & 923 \\
            & 16-bit & 1487 & 1151 & 939 \\ 
\bottomrule
\end{tabular}
\caption{Salient neurons under different thresholds (Thr.) across models.}
\label{table:salient_neurons_thresholds}
\end{table}

\begin{table}
\centering
\footnotesize
\setlength{\tabcolsep}{4pt}
\renewcommand{\arraystretch}{1.1}
\begin{tabular}{l l c c c}
\toprule
\textbf{Model} & \textbf{Quant.} & \textbf{0.1 (\%)} & \textbf{0.05 (\%)} & \textbf{0.01 (\%)} \\
\midrule
Phi-2 & 4-bit  & 7.16 & 7.13 & 0.02 \\
      & 8-bit  & 7.18 & 7.14 & 0.00 \\
      & 16-bit & 7.17 & 7.14 & 0.00 \\ \midrule
Llama-7B & 4-bit  & 0.02 & 0.00 & 0.00 \\
          & 8-bit  & 0.01 & 0.00 & 0.00 \\
          & 16-bit & 0.02 & 0.00 & 0.00 \\ \midrule
Qwen-3B & 4-bit  & 0.00 & 0.00 & 0.00 \\
         & 8-bit  & 0.00 & 0.00 & 0.00 \\
         & 16-bit & 0.00 & 0.00 & 0.00 \\ \midrule
Qwen-7B & 4-bit  & 0.00 & 0.00 & 0.00 \\
         & 8-bit  & 0.00 & 0.00 & 0.00 \\
         & 16-bit & 0.00 & 0.00 & 0.00 \\ \midrule
Mistral-7B & 4-bit  & 0.01 & 0.00 & 0.00 \\
            & 8-bit  & 0.00 & 0.00 & 0.00 \\
            & 16-bit & 0.01 & 0.00 & 0.00 \\
\bottomrule
\end{tabular}
\caption{Percentage of dead neurons under different activation thresholds across models.}
\label{table:dead_neurons_thresholds}
\end{table}

\end{document}